\documentclass[lettersize,journal]{IEEEtran}
\usepackage{amsmath,amsfonts}
\usepackage{algorithmic}
\usepackage{algorithm}
\usepackage{array}
\usepackage[caption=false,font=normalsize,labelfont=sf,textfont=sf]{subfig}
\usepackage{textcomp}
\usepackage{stfloats}
\usepackage{url}
\usepackage{verbatim}
\usepackage{graphicx}
\usepackage{cite}
\usepackage{hyperref}
\usepackage{academicons}
\usepackage{xcolor}
\usepackage{xspace}
\usepackage{tcolorbox}
\usepackage{multirow}

\usepackage{caption}
\newcommand{\orcid}[1]{\href{https://orcid.org/#1}{\textcolor[HTML]{A6CE39}{\aiOrcid}}}

\newcommand{\EnrichEvent}{\emph{EnrichEvent}\xspace}
\newcommand{\NamedEntities}{\emph{Named Entities}\xspace}
\newcommand{\Hashtags}{\emph{Hashtags}\xspace}
\newcommand{\Consolidation}{\emph{Consolidation}\xspace}
\newcommand{\Discrimination}{\emph{Discrimination}\xspace}

\newcommand{\TrendingDataExtraction}{\emph{Trending Data Extraction}\xspace}
\newcommand{\ContextualKnowledgeEnhancement}{\emph{Contextual Knowledge Enhancement}\xspace}
\newcommand{\EventClustering}{\emph{Event Clustering}\xspace}
\newcommand{\EventChainFormation}{\emph{Event Chain Formation}\xspace}
\newcommand{\EventSummarization}{\emph{Event Summarization}\xspace}

\usepackage{array}
\newcolumntype{P}[1]{>{\centering\arraybackslash}p{#1}}
\usepackage{tabularray}

\begin{document}

\title{\EnrichEvent: Enriching Social Data with Contextual Information for Emerging Event Extraction}

\author{Mohammadali Sefidi Esfahani (\href{https://orcid.org/0009-0007-0285-1545}{ORCID}), 
        Mohammad Akbari (\href{https://orcid.org/0000-0002-3321-5775}{ORCID}) \\
        \textit{Department of Mathematics and Computer Science,} \\
        \textit{Amirkabir University of Technology, Tehran, Iran} \\
        Email: \{mohammadali.esfahani99@gmail.com, akbari.ma@aut.ac.ir\}%
}

\markboth{Arxiv, June 2025}%
{Shell \MakeLowercase{\textit{et al.}}: A Sample Article Using IEEEtran.cls for IEEE Journals}

\maketitle

\begin{abstract}
Social platforms have emerged as crucial platforms for distributing information and discussing social events, offering researchers an excellent opportunity to design and implement novel event detection frameworks. 
Identification of \textit{unspecified events}, detection of events without using any prior knowledge, enables governments, aid agencies, and experts to respond swiftly and effectively to unfolding situations, such as natural disasters, by assessing severity and optimizing aid delivery. Additionally, it provides valuable insights for marketing, commerce, and financial markets.
Social data is characterized by misspellings, incompleteness, word sense ambiguation, and irregular language. While discussing an ongoing event, users share different opinions and perspectives based on their prior experience, background, and knowledge. 
Prior works primarily leverage tweets' lexical and structural patterns to capture users' opinions and views about events. However, tweets' lexical and structural aspects do not perfectly reflect their views. 
Additionally, extracting discriminative features and patterns for evolving events by exploiting the limited structural knowledge is almost infeasible. 
In this study, we propose an end-to-end novel framework, \EnrichEvent, to identify unspecified events from streaming social data.
In addition to lexical and structural patterns, we leverage contextual knowledge of the tweets to enrich their representation and gain a better perspective on users' opinions about events. 
Compared to our baselines, the \EnrichEvent framework achieves the highest values for \Consolidation outcome with an average of 87\% vs. 67\% and the lowest values for \Discrimination outcome with an average of 10\% vs. 16\%.
These results show that the \EnrichEvent framework understands events well and distinguishes them perfectly.
Moreover, the \TrendingDataExtraction module in the \EnrichEvent framework improves efficiency by reducing \textit{Runtime} by up to 50\%.
It achieves this by identifying and discarding irrelevant tweets within message blocks.
As a result, the \EnrichEvent framework becomes highly scalable for processing streaming data.
Our source code and dataset are available in our official replication package~\cite{replication2023EnrichEvent}.

\end{abstract}


\section{Introduction} \label{sec:introduction}

\IEEEPARstart{s}{\lowercase{ocial}} media platforms have become essential in our daily lives by enabling us to communicate with friends, share information, exchange ideas, and acquire knowledge~\cite{Chen2013EmergingTD}. 
Worldwide, billions of people actively use social media, and experts predict that active users will grow from 2.86 billion in 2017 to 4.41 billion in 2025~\cite{SocialMedia_Statistics}. 
Consequently, social media now plays a crucial role in discussing and disseminating information about important events~\cite{thapa2024stance}. 
When social incidents occur, experts and ordinary users engage in multifaceted discussions, as seen in Twitter conversations during events like the \textit{Notre Dame Cathedral fire}~\cite{Aldhaheri_2017, Sun_Xiang_Wu_2015}. 
Users' active participation and real-time updates transform these platforms into primary sources for identifying, analyzing, and investigating social events, offering a unique chance to detect newsworthy events before traditional media~\cite{parekh2024event}. 
Therefore, academic researchers and industry professionals are motivated to develop frameworks to uncover social events and capture real-time discussions.

Traditionally, an \emph{event} is defined as an important happening at a specific time and place~\cite{allan1998topic}. 
With the rise of social media, this definition transforms to the occurrences that attract sudden bursts of public attention while still being tied to a specific time and place~\cite{boettcher2012eventradar}. 
Following prior work\cite{RealTime_EventDetection_Twitter}, we consider each event as a single chain of clusters containing closely related entities in time and semantic dimensions.
When an event unfolds, governments, aid agencies, journalists, and experts urgently require swift and precise identification to respond effectively~\cite{Governments, huang2021similarity, upadhyay2024satcobilstm}. 
For example, during natural disasters, relief organizations rely on event detection to quickly assess factors like incident severity, victim numbers, and the optimal delivery of aid~\cite{nugent2017comparison,pekar2020early,sreenivasulu2020comparative}. 
Moreover, timely awareness of events is crucial for marketing, commerce, and financial markets, where understanding trends can give market leaders a competitive edge~\cite{Mohan, Shah}. 
Despite its significance, the rapid and accurate identification of unforeseen social events still faces substantial challenges.

1)~\textbf{Streaming and Evolutionary}. 
Social media services provide a streaming source of information in which people freely create content around real-world events. While working on streaming sources, processing the incoming data in a single stage, and modeling the evolving characteristics of an event is challenging. Moreover, transforming a simple event into a trending social topic, which everyone discusses and shares their opinions about on social platforms, is a complex and evolving phenomenon consisting of three main steps. \textbf{a)} An event happens in the real world. \textbf{b)} Observers, journalists, and ordinary people start a discussion about the ongoing event on social platforms. \textbf{c)} The event turns into a hot event for a while and fades out after a period~\cite{Chen2013EmergingTD}.

2) \textbf{Variation in Language Expressions and Aspects of Opinions}. 
Twitter users produce and consume information in a very informal and irregular way, and the published tweets usually include idioms, abbreviations, misspellings, irregular language, and emojis. Due to the vast and diverse user base, Twitter usually contains different word sequences describing the same idea. Additionally, in most languages, many words have various meanings based on the context. For example, the word \textit{fair} has multiple meanings of \textit{carnival}, \textit{treating someone right}, and \textit{having light skin/hair} based on the context. While discussing an event, users share different opinions and perspectives based on their prior experience, background, and knowledge. The variation in users' language and opinions makes detecting social events challenging for event detection frameworks.

Prior works~\cite{Bursty_Event_Detection, Yang2018AnED, RealTime_EventDetection_Twitter} primarily leverage tweets' lexical and structural patterns to capture users' opinions and views about events. 
However, various users discuss ongoing events on Twitter, and the lexical and structural aspects of tweets do not perfectly reflect their opinions. 
In this paper, we design an end-to-end novel framework, \EnrichEvent, to identify unspecified events from streaming data.
In addition to lexical and structural patterns, we leverage contextual knowledge of the tweets to enrich their representation and gain a better perspective on users' opinions. 
Exploiting contextual knowledge assists the model in understanding the existing relationship among tweets and generating a high-quality representation of users' opinions. 

Following prior work~\cite{RealTime_EventDetection_Twitter}, we examine the performance of the \EnrichEvent framework with the outcomes of \Consolidation and \Discrimination. The high values for \Consolidation outcome ensure that all the clusters in event chains refer to the
same event. The \Discrimination outcome, on the other hand, quantifies the ability of frameworks to distinguish the events from each other. To evaluate the efficacy of the event detection frameworks, we run simulations on streaming data from Twitter for over two weeks.

\subsection{Research Questions}

We present our research questions and briefly summarize our findings below.

\noindent\textbf{RQ1, Replication \& Evaluation: How well does the \EnrichEvent framework perform compared to the baselines?}

In the first research question, we replicate the baselines~\cite{word2vec, BERT, Yang2018AnED, RealTime_EventDetection_Twitter} and compare their performance with our novel framework, \EnrichEvent. The main novelty of the \EnrichEvent framework is that it integrates the lexical, structural, and contextual knowledge of tweets to represent users' opinions about ongoing events better.

\textbf{Result summary}: 
Our simulations show that the \EnrichEvent framework achieves the highest \Consolidation with the average of 87.4\% and the lowest \Discrimination with the average of 10\% among the studied baselines. These results validate that integrating the contextual, structural, and lexical knowledge in tweets is crucial for capturing the diverse perspectives of users and ensuring robust event detection. 

\noindent\textbf{RQ2, Representation Enhancement: How does leveraging the contextual knowledge along with the lexical and structural aspects enhance the quality of event detection?}

Compared to prior works~\cite{Yang2018AnED, RealTime_EventDetection_Twitter}, we target the contextual knowledge of tweets to enrich the representation of users' opinions and views about the ongoing tweets. In this research question, we examine the quality of generated clusters in the \EventClustering module based on the \textit{Fraction of Relevant Entities (FRE)}, \textit{Davies-Bouldin}, \textit{Calinski-Harabasz}, and \textit{Silhouette} scores.

\textbf{Result summary}: 
Compared to baselines, the \EnrichEvent achieves the highest \textit{Fraction of Relevant Entities (FRE)} with an average of 93\% across all the generated clusters. Moreover, the investigation of clustering metrics confirms that leveraging contextual knowledge yields better representation and higher-quality clustering.

\noindent\textbf{RQ3, Impact of Trending Data Extraction Module: What is the impact of filtering out the pointless tweets in message blocks on the performance of the \EnrichEvent framework?}

Since the unspecified event detection frameworks deal with vast stream data, their scalability should be high. In the \EnrichEvent framework, the \TrendingDataExtraction module aims to filter out the pointless tweets from message blocks to increase the scalability of the \EnrichEvent framework. In this research question, we examine the impact of the \TrendingDataExtraction component on the performance and \textit{Runtime} of the \EnrichEvent framework.

\textbf{Results summary}: 
Our experiments show that the \TrendingDataExtraction module reduces the \textit{Runtime} of the \EnrichEvent framework up to 50\% while maintaining high-quality event detection.

This paper is organized as follows: In the next section, we briefly survey existing event detection frameworks. In Section~\ref{sec:problem}, we define the problem of unspecified event detection. In Section~\ref{sec:methodology} and~\ref{sec:results}, we present our methodology and report results for each research question. In Section~\ref{sec:discussion}, we discuss our key findings in this study and the limitations of our work. Finally, we conclude our study in Section~\ref{sec:conclusion}.

\begin{figure*}
  \includegraphics[width=\textwidth]{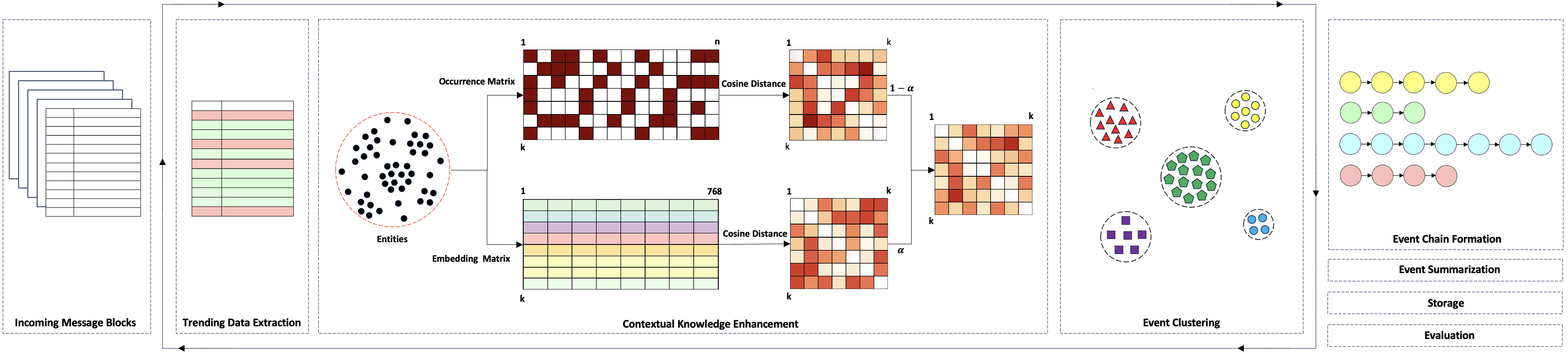}
  \caption{An overview of the functionality of our proposed framework, \EnrichEvent. The framework processes message blocks in batches corresponding to the incoming social stream. First, the \TrendingDataExtraction module filters out pointless tweets. Next, the \ContextualKnowledgeEnhancement component extracts \NamedEntities and \Hashtags from the tweets and builds their representation matrices. Then, the \EventClustering module categorizes the extracted entities into clusters based on their lexical, structural, and contextual representation. Finally, the \EventChainFormation module connects related clusters across consecutive message blocks to form event chains, and the \textit{Storage} component saves each event's information as a JSON file in the database.}
  \label{fig:overview}
\end{figure*}

\section{Related Works} \label{sec:related}
Since Twitter introduced the Twitter API and researchers can easily collect real-time data, Event detection from Twitter has received noticeable attention in recent years~\cite{Survey_2016, ED_SWE}. We can divide the problem of social event detection into three sub-categories. Identification of \textit{predetermined events}, \textit{sub-events}, and \textit{unspecified events}~\cite{atefeh2015survey}. Our work belongs to the last category, and in this section, we investigate these three sub-categories and explain some related works that motivate us to work in this field.

\subsection{Identification of predetermined events}
In the first category, researchers usually leverage datasets such as \textit{Event2012}~\cite{Event2012} and \textit{Event2018}~\cite{Event2018}, which contain predefined events in different event classes. Two critical challenges exist in this category, one of which is providing researchers with opportunities to design novel frameworks. 

\textbf{1) Scalability}. 
Most of the recent papers convert the message blocks to message graphs to model the dependencies between the tweets. Moreover, they leverage the GNN models to generate the embedding of the tweets~\cite{cao2021knowledge, peng2021streaming, peng2022reinforced, qiu2024heterogeneous}. Although utilizing message graphs and GNN models enhances the messages' embedding, they reduce the scalability of the frameworks. 

\textbf{2) Usability}.
The proposed frameworks should be trained on labelled datasets, which makes their use in real-world problems challenging.~\textit{Ren et al.}~\cite{Qsgnn} investigate this challenge and design a novel framework, \textit{QS-GNN}. While training their proposed model, it only requires the labelled messages in the first time frame. Although they minimize the model's dependence on labelled data, leveraging these frameworks for real-world problems is still challenging.

\subsection{Identification of sub-events}
In this category, researchers consider a specific event, e.g., a football match, and aim to identify the relevant sub-events. Designing a decent approach to find the correlations between the tweets and a given event is the main challenge of this category. \textit{Bekoulis et al.}~\cite{bekoulis-sub} investigate the temporal aspects of the tweets and propose a two-stage framework that detects the sub-events by analyzing their evolution over time. Moreover, ~\textit{Lu et al.}~\cite{lu2021hashtag} focus on semantics to enrich the representation of tweets. In addition, they use hashtags and their n-grams to capture the correlations between tweets suitably. Though much research has been done in this category, researchers can enhance the existing frameworks differently.

\subsection{Identification of unspecified events}
While working on this category, there needs to be an antecedent description of the events, and researchers aim to detect general events. Researchers typically investigate different aspects of the tweets in this category to discover their existing relations. As mentioned in the previous section, processing streaming data is a challenging task, and \textit{Comito et al.}~\cite{Bursty_Event_Detection} tries to design an event detection framework that is robust to processing issues. Moreover, they exploit the semantics in the tweets by working on the 2-grams of words, hashtags, and mentioned users. Eventually, they identify the events by grouping the related tweets using the incremental clustering approach. 

The main idea of~\textit{Yang et al.}~\cite{Yang2018AnED} is to cluster similar hashtags together, and clusters of hashtags represent occurring events. They focus on the occurrence of hashtags in tweets and consider hashtags as a bag of words and a bag of hashtags. Eventually, they use the concatenation of hashtag-hashtag and hashtag-words co-occurrence vectors as a final feature vector for clustering. The main challenge of this work is that users may not necessarily use hashtags in all tweets. Besides, posted hashtags in tweets are not necessarily relevant to their content. 

~\textit{Fedoryszak et al.}~\cite{RealTime_EventDetection_Twitter} considers happening events as chains of clusters. They build a real-time system to identify groups of event-related entities. Then, they link these clusters together and generate cluster chains that represent the events. In this study, the authors only focus on the lexical representation of \NamedEntities and use occurrence vectors to cluster them.

Due to budget constraints on the Twitter API, we could not access benchmark datasets such as \textit{Event2012}~\cite{Event2012} and \textit{Event2018}~\cite{Event2018}. 
Consequently, we decided to work on identifying the unspecified events from Twitter.  
Prior works~\cite{Bursty_Event_Detection, Yang2018AnED, RealTime_EventDetection_Twitter} primarily leverage users' lexical and structural patterns to capture their opinions about events. 
However, users exhibit varied characteristics and discuss ongoing social events from different perspectives based on their backgrounds and experiences. 
In this study, we target the contextual representation of tweets and propose a novel event detection framework, \EnrichEvent, which accommodates users' diverse characteristics and language variations.

\section{Problem Statement} \label{sec:problem}
In this section, we present the key terms of this study and formally define the problem of unspecified event detection from social media data.

\noindent \textbf{Definition 2.1.} 
Social stream $\mathcal{S}=\{\mathcal{M}_1, \ldots, \mathcal{M}_{i-1}, \mathcal{M}_i, \ldots\}$ is defined as a consecutive sequence of message blocks. We denote message block $\mathcal{M}_i$ as $\mathcal{M}_i=\{m_j | 1 \le j \le |\mathcal{M}_i| \}$, where $|\mathcal{M}_i|$ is total numbers of the tweets in time period $[t_i, t_{i_1})$ and $m_j$ denotes a single tweet.

\noindent\textbf{Definition 2.2.} 
We can group all the tweets in message block $\mathcal{M}_i$ to finite set of clusters $\mathcal{C}_i=\{c_j | 1 \le j \le |\mathcal{C}_i| \}$, where messages in each cluster are contextually related.

\noindent\textbf{Definition 2.3.} 
Following prior work~\cite{RealTime_EventDetection_Twitter}, we define a social \emph{event} $e=\{c_{ij} | 1 \le i \le |\mathcal{M}_i|, 1 \le j \le |\mathcal{C}_i| \}$ as a continuous sequence of contextually related clusters where all the clusters refer to the same event.

\noindent\textbf{Objective:} 
Given a social stream of message blocks $\mathcal{S}$, we aim to design and implement an end-to-end framework, $F$, that handles the streaming nature of social media data and detects unspecified social events, $E$, from the incoming social streams.

\begin{equation}
    \centering
    F: \mathcal{S} \rightarrow E=\{e_1, e_2, \ldots\}
\end{equation}

\section{Methodology} \label{sec:methodology}


Since events start at a specific time point and extend over multiple time windows, in this study, we focus on the need for an end-to-end pipeline that handles the streaming nature of social media data and detects unspecified events by processing incoming social streams. 
Our proposed framework, \EnrichEvent, comprises seven components: \textit{\TrendingDataExtraction}, \textit{\ContextualKnowledgeEnhancement}, \EventClustering, \EventChainFormation, \textit{\EventSummarization}, and \textit{Storage \& Evaluation}.
In this section, we first go through each component of the \EnrichEvent framework and explain its roles and functionality. 

\subsection{Trending Data Extraction}

Social media data is often known as a noisy and sparse information source. Prior works~\cite{karimi2023enhancement, li2012twevent} highlight the need to filter irrelevant and pointless data while working on social data. Morabia et al.~\cite{SEDTW} report that users publish around 500 million daily tweets. However, nearly 40\% of these tweets are just \emph{pointless babble}, which are insignificant to the event detection task. 
In this study, we focus on detecting unspecified events that nobody knows about. Since there is no predefined data about the occurring events in this category, the proposed pipelines for unspecified event detection deal with extensive data samples.
We should filter out the pointless tweets from the message blocks to make the \EnrichEvent framework scalable. In this way, \EnrichEvent processes streaming social data and extracts social events in an almost online setting.
To achieve this goal, we place the \TrendingDataExtraction component as the first stage of our framework.
This module examines the tweets in the incoming message blocks to classify the ones that include newsworthy information that potentially refers to an ongoing event. 

Inspired by prior studies~\cite{Chen2013EmergingTD, chen2015event}, we embed a neural network in the \TrendingDataExtraction component to estimate the probability of referring to an event for each tweet in the message block.
The proposed architecture of this neural network includes two primary segments.
\textbf{1) Feature Extraction.} In the feature extraction stage, we implement a convolutional neural network (CNN) to learn rich representations from the input tweet. We convert the input text into dense vector representations using a BERT-based embedding layer to capture contextual meaning. Then, multiple convolutional layers apply filters to detect important local patterns in the text. Max-pooling layers are used to retain the most informative patterns while reducing dimensionality. This process ensures that the network focuses on meaningful parts of tweets before estimating the probability of referring to an event.
\textbf{2) Classification.} In the classification stage, we flatten the output from the feature extraction segment and feed it through a fully connected neural network. The dense layers learn complex relationships between the extracted features, while non-linear activation functions enhance the model's ability to capture intricate patterns. Finally, a sigmoid activation function produces the probabilities of predicted classes, determining the most likely category for the input text. 
By combining CNN-based feature extraction with a fully connected classifier, the \TrendingDataExtraction module effectively captures local and high-level text representations, leading to high accuracy in filtering pointless tweets.

We adopt a supervised setting to train the neural network used in the \TrendingDataExtraction component, $\mathcal{T}$.
This neural network calculates the probability that a tweet refers to a potential event by solving the optimization problem described in Equation~\ref{eq:trendDetection}. This model takes extracted features from tweets as input vectors, $\mathbf{X} \in \mathbb{R}^{n \times d}$, and computes the probability of a tweet either containing general non-informative content (negative class) or referring to a potential event (positive class). $\mathcal{L}(\cdot)$ is the binary cross-entropy loss function in this equation. Matrix $\mathbf{W} \in \mathbb{R}^{d \times h}$ contains all trainable parameters that the neural network optimizes during the training process. Label vector $\mathbf{Y} \in \mathbb{R}^{n \times 1}$ indicates whether each tweet in the given message block refers to an event (label 1) or not (label 0).

\begin{equation}
    \label{eq:trendDetection}
    \centering
    \arg \min_\mathbf{W} \mathcal{L}(\mathbf{X}, \mathbf{W}, \mathbf{Y}),
\end{equation}

The classifier that we implement and train for the \TrendingDataExtraction component gets the tweets in a given message block as input and identifies tweets with a high probability of referring to an event. This classifier allows us to filter out the pointless tweets based on the minimum threshold for the likelihood of referring to a potential event.
Therefore, the message blocks come through the \TrendingDataExtraction component at the first stage, and the classifier filters out the pointless tweets based on the threshold $\sigma$.
Ultimately, the \EnrichEvent framework passes the filtered message blocks to the next component for representation extraction.

\subsection{Contextual Knowledge Enhancement}

When users debate ongoing events on social platforms, they discuss different aspects and have diverse opinions based on their prior background and experience. 
Working on lexical and structural aspects of tweets is a conventional way to design event detection models~\cite{zhang2016improving, ananthakrishnan2007automatic, RealTime_EventDetection_Twitter}. 
However, designing a well-performing framework that covers different aspects and opinions is challenging because of the users' diverse characteristics and language variations.
After filtering the pointless and general tweets, we implement the \ContextualKnowledgeEnhancement component to extract a rich representation from the newsworthy tweets for event detection.
This component aims to get the filtered message blocks and generate an expressive representation for them, covering structural and semantic aspects of tweets.

In the \ContextualKnowledgeEnhancement component, we target two main types of entities that users mention in their tweets, including \NamedEntities and \Hashtags, to extract a rich representation from tweets.
\NamedEntities are critical indicators for detecting events since an event happens at a specific time and location and includes descriptive actors. Users use \NamedEntities to demonstrate and discuss one of these attributes related to an event. 
\Hashtags, on the other hand, are intrinsically incorporated into social posts to reveal the correlation of the tweets to an event or a topic, making them invaluable indicators of tweets referring to an event. 
For instance, in the tweet' \textit{\underline{Silicon Valley} Bank of the \underline{United States} collapsed! The largest bank \underline{\#bankruptcy} in the \underline{USA} since the crisis of \underline{2008} !!!}' a user leverages the \NamedEntities and \Hashtags to discuss a social event. 
In this tweet, the \NamedEntities indicate that the attributes of the event are \textit{Silicon Valley}, \textit{USA}, \textit{United States}, and \textit{2008}. In addition, \textit{\#bankruptcy} demonstrates that this social event corresponds to bankruptcy.
In this study, we consider \NamedEntities and \Hashtags as the representatives of tweets in the incoming message block.
We embed BERT's NER model and regular expressions in this component to retrieve the posted \NamedEntities and \Hashtags from the incoming filtered message block.
Table~\ref{tab:entity} contains the possible \NamedEntities that users may mention in their tweets to describe their opinion about an event.

\begin{table}
  \centering
  \small
  \scriptsize 
  \caption{The BERT NER model supports various types of Named Entities. Users leverage these Named Entities to express their opinions and perspectives about the events.}
  \begin{tabular}{|p{1.75cm}|p{6cm}|} 
    \hline
    \textbf{Named Entity} & \textbf{Description} \\ 
    \hline
    Person & Names of people, including fictional characters. \\ 
    \hline
    ORG & Organizations, including companies, institutions, etc. \\ 
    \hline
    LOC & Locations that are not geopolitical entities  \\ 
    \hline
    GPE & Geopolitical entities, including countries, cities, and states. \\ 
    \hline
    Date & Specific dates or time expressions  \\ 
    \hline
    Time & Specific times of the day \\ 
    \hline
    Money & Monetary values, including currency symbols \\ 
    \hline
    Percent & Percentage expressions  \\ 
    \hline
    Quantity & Measurements  \\ 
    \hline
    Event & Mentioned events  \\ 
    \hline
    Product & Names of products, including cars, software, or devices. \\ 
    \hline
    Works of Art & Titles of books, songs, movies, or paintings. \\ 
    \hline
    Law & Names of laws, regulations, or treaties \\ 
    \hline
  \end{tabular}
  \label{tab:entity}
\end{table}

Assume that users mention $k$ unique entities in an incoming message block $\mathcal{M}_i$ which contains $|\mathcal{M}_i|$ tweets. After retrieving the posted entities from the incoming message block, the \ContextualKnowledgeEnhancement component builds two representation matrices for each entity. 
The first is the \textit{Occurrence Matrix $\mathbf{O} \in \mathbb{R}^{k \times |\mathcal{M}_i|}$} that represents entities' lexical and structural aspects.
We consider each entity $k_i$ as a bag of tweets to construct the \textit{Occurrence Matrix}. 
In this matrix, element $e_{ij}$ shows the frequency of entity $k_i$ in tweet $t_j$ and the $i^{th}$ row demonstrates the lexical and structural representation of entity $k_i$

The second representation matrix, \textit{Embedding Matrix $\mathbf{E} \in \mathbb{R}^{k \times 768}$}, captures the contextual representations of the extracted entities. 
To compute the embedding vector for each entity \(k_i\), we combine tweets from two sources: the top retweeted tweets and a seeded random selection of tweets where \(k_i\) appears. 
We introduce a hyperparameter, $\beta$, to control the number of chosen tweets per entity. 
Specifically, we choose $\frac{\beta}{2}$ tweets from the top retweeted group and randomly (seeded) choose the remaining tweets from the other tweets in which \(k_i\) appears. 
Next, we concatenate the selected tweets and use a BERT language model~\cite{ParsBERT} to generate the final embedding vector. 
Finally, we repeat this procedure for every entity to construct an embedding matrix, where the \(i^{th}\) row represents the contextual embedding of the \(i^{th}\) entity.

In our simulations, the goal is to compare the effectiveness of event detection frameworks directly. 
When the \ContextualKnowledgeEnhancement module selects tweets for building the \textit{embedding matrix}, each framework selects the same tweets, eliminating any variation from randomly selecting different tweets. By consistently choosing the same seeded random tweets, the variation in evaluation metrics will only result from the different frameworks. 
This seeded random approach achieves our goal of evaluating different event detection frameworks and explicitly does not evaluate how randomly selected tweets would have impacted the outcome measures.

We convert the \textit{Occurrence Matrix} and the \textit{Embedding Matrix} into separate entity-to-entity distance matrices using the \textit{Cosine} distance function. This yields two matrices that quantify the cosine distance between every pair of entities.
Next, we aggregate these distance matrices into a single distance matrix by taking a weighted average, where we assign a weight of $1-\alpha$ to the \textit{Occurrence Matrix} and $\alpha$ to the \textit{Embedding Matrix}. The $\alpha$ gives the flexibility to users of the \EnrichEvent framework to adjust the importance of the representation matrices.
At the end of this stage, the \ContextualKnowledgeEnhancement component produces a final entity-to-entity distance matrix and passes it to the \EventClustering component for clustering related entities. 
This final distance matrix captures the lexical, structural, and contextual aspects of users' opinions about the events in the incoming message block.

\subsection{Event Clustering}

Users have unique backgrounds, experiences, and personal perspectives about the events and interpret and discuss them using various attributes and entities.
The entity-to-entity distance matrix from the previous stage comes to the \EventClustering component, which aims to cluster the related entities that refer to the same event in a single cluster.
Among the existing clustering models, we embed the HDBSCAN~\cite{HDBSCAN} model in this component because it does not require a predefined number of clusters and can dynamically adapt to the varying structure of streaming data. 
Its ability to identify clusters of different densities while filtering out noise makes it particularly suitable for grouping entities representing the same event from diverse perspectives.
Given an entity-to-entity distance matrix as input, the HDBSCAN model clusters the extracted entities based on their lexical, structural, and contextual representation.
At the end of this stage, the \EventClustering module generates a group of clusters from the incoming message block to the pipeline.
In the next step, the \EventChainFormation component links the extracted clusters to the clusters in the previous time window to build the event chains.

\subsection{Event Chain Formation}

Since events evolve continuously over time, we cannot identify unspecified events through static analysis; instead, we track their evolution across consecutive message blocks. 
The goal of the \EventChainFormation module is to link clusters from different time windows to form coherent event chains. 

Let $\mathcal{C}_i= \{c_1, c_2, \ldots, c_m\}$ denote the set of clusters generated from message block $\mathcal{M}_i$ by the \EventClustering component. 
The \EventChainFormation module constructs a bipartite graph by connecting clusters in $\mathcal{C}_i$ with those in the previous time window, $\mathcal{C}_{i-1}$, where clusters serve as nodes. 
Furthermore, it adds an edge between a cluster $c_k \in \mathcal{C}_i$ and a cluster $c_l \in \mathcal{C}_{i-1}$ if they share more than a minimum threshold $\delta$ of common entities, the weight of the edge equals the number of these shared entities. 
Then, the \EventChainFormation module applies the Hungarian algorithm~\cite{kuhn1955hungarian} to perform a maximum weighted matching, linking clusters from the current message block $\mathcal{M}_i$ with those from the previous block $\mathcal{M}_{i-1}$. 
When a cluster is successfully linked, the module transfers the identifier from the previous cluster to the current state. If a cluster remains unlinked, it shows that the users stop discussing that event, and the module assigns it a new unique event ID. 
Finally, the \EventChainFormation module sends the detected events to the \EventSummarization module to generate event summaries.

\subsection{Event Summarization}

While feeding the daily message blocks to the \EnrichEvent pipeline, the \EventChainFormation component efficiently detects events by representing them as cluster chains.
For each detected event, \(e\), there exists a set of corresponding tweets, \(T=\{t_1, t_2, \ldots, t_n\}\), where users share diverse perspectives on that event.
Once the \EnrichEvent framework identifies an event and confirms that its cluster chain has stabilized, the associated tweets are forwarded to the \EventSummarization module.
This module utilizes the GPT-3.5 language model to generate concise and informative event summaries based on corresponding tweets.
Thus, for each detected event, the \EventSummarization component produces an event summary and passes it to the next stage for storing the results.

\subsection{Storage \& Evaluation}

At the end of the \EnrichEvent pipeline, the detected events and their corresponding event summary from the \EventSummarization module feed into the \textit{Storage} component.
The \textit{Storage} module extracts informative information about the events, such as event ID, event period, event summary, details of corresponding entities/hashtags/usernames to each event, and tweets with the highest like/reply/retweet count.
Finally, the \textit{Storage} module stores details of each identified event as a JSON file in the database.
More details about the attributes of the final output of the \EnrichEvent framework are available in Appendix~\ref{Appendix:output}.

Furthermore, the \EnrichEvent pipeline includes an \textit{Evaluation} component that examines the framework's performance and the results from various perspectives. We discuss the results and evaluation details in Section~\ref{sec:exSetup} and~\ref{sec:results}.

\subsection{Framework Overview}

In this study, we focus on identifying unspecified events from Twitter. 
The \EnrichEvent framework continuously processes the social stream to detect ongoing events discussed by users. 
A key advantage of our approach is its ability to extract social events from streaming data in real time. 
Figure~\ref{fig:overview} illustrates the architecture and functionality of the \EnrichEvent framework. 
This section reviews the \EnrichEvent pipeline and explains how it detects events from the social stream. 

Social stream $\mathcal{S}$ enters the pipeline as daily message blocks, denoted by $\mathcal{M}_{i}$. 
The \TrendingDataExtraction module first filters out irrelevant tweets and retains those likely to refer to an event. 
Next, the \ContextualKnowledgeEnhancement component extracts entities from tweets and constructs an entity-to-entity distance matrix based on lexical, structural, and contextual features. 
The \EventClustering module then groups related entities into clusters based on their computed distances. 
Subsequently, the \EventChainFormation module connects clusters from the current message block $\mathcal{M}_{i}$ with clusters from the previous block $\mathcal{M}_{i-1}$, forming evolving chains that represent ongoing events. 
Once an event is detected, the \EventSummarization component generates a summary that captures users' opinions and passes that to the \textit{Storage} component. 
Finally, the \textit{Storage} module saves the detailed information of each detected event as a JSON file in the database.

\section{Experimental Setup} \label{sec:exSetup}
This section presents our baselines, dataset, evaluation outcomes, and the implementation setting used to run our simulations.

\subsection{Baselines}

We compare the \EnrichEvent framework against base models and prior event detection methods. Since these baselines generate different entity representations, we adjust the \ContextualKnowledgeEnhancement component for each baseline to ensure a fair comparison. All baselines are implemented under the same experimental setting.
The base models that we replicate in this study are \textit{Word2vec}~\cite{word2vec} and \textit{BERT}~\cite{BERT}. These models generate embedding vectors of 768 and 64 dimensions per entity, respectively. 
To implement these baselines, we utilize the contextual representation of entities within the \ContextualKnowledgeEnhancement component to construct the entity-to-entity distance matrix, without incorporating structural or lexical information.

In addition to base models, we investigate the performance of event detection frameworks from prior research and compare it with the \EnrichEvent framework.
\textit{Fedoryszak et al.}~\cite{RealTime_EventDetection_Twitter} propose an approach that constructs an interaction graph for each message block, capturing relationships among entities based on their co-occurrence in tweets. They apply \textit{Louvain} community detection to these graphs and build cluster chains across consecutive message blocks to identify events.
\textit{Yang et al.}~\cite{Yang2018AnED} take a hashtag-centered approach, considering hashtags as key indicators of users' opinions and perspectives on events. Their method generates co-occurrence vectors for hashtag-hashtag and hashtag-word relationships and concatenates them to form entity representations.

\subsection{Datasets}

\textbf{Trend detection dataset}. 
We collect an offline trend detection dataset from Twitter to train the classifier in the \TrendingDataExtraction component. 
The dataset consists of 1.6 million Persian tweets spanning various categories, including sports, technology, art, etc. This diversity enhances the generalizability and effectiveness of the \TrendingDataExtraction component. 
The dataset contains key attributes such as \textit{timestamp}, \textit{tweet ID}, \textit{text}, \textit{category}, \textit{username}, \textit{reply count}, \textit{like count}, and \textit{retweet count}. 
For labeling, we define a set of key phrases for each category and employ an automated approach to label each tweet based on the presence of these key phrases.
Finally, we sort the tweets of each category in chronological order and use 80\% of the tweets for training and 20\% for evaluation.

\textbf{Event detection dataset}.
In addition to the trend detection dataset, we construct another dataset to conduct experiments and evaluate the overall performance of \EnrichEvent on Twitter streaming data.   
We collect Persian tweets over two weeks, from 2023/2/28 to 2023/3/14, resulting in approximately 76k tweets. These tweets are segmented into daily message blocks for analysis.
We manually identify occurring events within the dataset to assess the framework's effectiveness. For this, we sample 500 tweets from each message block based on token frequency, ensuring that the sampled tweets reflect the actual distribution of words in the dataset. Tweets containing more frequently occurring words have a higher probability of being selected.  
Further details on the collected dataset can be found in Appendix~\ref{Appendix:data}.  

\subsection{Evaluation Metrics} \label{sec:metric}

While representing the events as a chain of entity clusters across consecutive time windows, evaluating the quality of each event chain and the separation of events from each other is critical.
Following prior work~\cite{RealTime_EventDetection_Twitter}, we measure the \Consolidation and \Discrimination metrics to assess the quality of event chains.
The \Consolidation outcome checks if entities in the same chain truly belong to the same event, and \Discrimination checks if entities from different events are kept apart.
By using both, we gain insights into whether each framework merges entities correctly for each event and avoids mixing entities from different events.
In this study, the $\big\uparrow$ next to the names of evaluation metrics shows that a higher value indicates better results, and the $\big\downarrow$ next to their names shows that a lower value of that metric leads to better results.

\textbf{Consolidation$\big\uparrow$.}
\Consolidation measures how well all entities in a chain are tied to the same event.
Let $T_e$ be the set of time windows where event $e$ occurs, $\mathcal{U}_t$ the total number of possible entity pairs in time window $t$, and $\mathcal{A}_t$ the number of those pairs that truly refer to event $e$. 
Equation~\ref{eq:consolidation} takes the fraction of correctly related entity pairs in each time window, then averages these fractions over all time windows for event $e$.
A higher value of $\mathcal{C}(e)$ indicates that entities belonging to event $e$ are consistently clustered together, reflecting better \Consolidation.

\begin{equation}
    \label{eq:consolidation}
    \mathcal{C}(e) = \frac{1}{|T_e|} \sum_{t \in T_e} \frac{\mathcal{A}_t}{\mathcal{U}_t}
\end{equation}

\textbf{Discrimination$\big\downarrow$}: 
\Discrimination measures how well our method avoids grouping unrelated entities together, i.e., entities that do not belong to the same event.
Let $\mathcal{B}_t$ denote the number of unrelated entity pairs incorrectly placed in the same cluster at time window $t$, and $\mathcal{U}_t$ is the total number of possible entity pairs.
Based on Equation~\ref{eq:discrimination}, $\mathcal{D}(e)$ calculates the fraction of unrelated pairs for each time window and then averages these fractions across all time windows for event $e$.
A lower value of $\mathcal{D}(e)$ means fewer unrelated entities are merged, indicating better \Discrimination between the events that users discuss.

\begin{equation}
    \label{eq:discrimination}
    \mathcal{D}(e) = \frac{1}{|T_e|} \sum_{t \in T_e} \frac{\mathcal{B}_t}{\mathcal{U}_t}
\end{equation}

An ideal unspecified event detection framework achieves high values for the \Consolidation outcome, and low values for the \Discrimination outcome. These results show that the framework can effectively distinguish events and their corresponding entities from each other.

\textbf{Fraction of Relevant Entities (FRE)$\big\uparrow$}: 
During the construction of cluster chains and event identification, the attending clusters in the event chains must include relevant entities to the events. 
To quantify the quality of entity clusters, we calculate the \textit{Fraction of Relevant Entities (FRE)} in all the entity clusters. 
This metric measures the proportion of entities in a cluster that are directly relevant to the occurring events, and a higher value indicates better cluster purity. 
Based on the Equation~\ref{eq:fre}, we calculate the \textit{FRE} outcome by dividing the number of relevant entities by the total number of entities in the cluster of event chains. 
\begin{equation}
\label{eq:fre}
\textit{FRE} = \frac{\text{NumRelevantEntities}}{\text{TotalNumEntities}}
\end{equation}

In this study, we extend the prior works~\cite{chen2015event, RealTime_EventDetection_Twitter} and focus on leveraging the contextual representation along with the structural and lexical representation of entities.
In addition to the proportion of relevant entities in event chains, we examine the quality of generated clusters from various perspectives, including similarity, variance, and distance of clusters.

\textbf{Silhouette Score$\big\uparrow$}~\cite{Silhouette}: 
The \textit{Silhouette Score} measures how similar an entity is to its own cluster compared to other clusters. In Equation~\ref{eq:sil}, \(a\) represents the average distance from an entity to all other entities in its cluster, while \(b\) is the minimum average distance from that entity to entities in a different cluster. Higher values for the silhouette score imply that the \EventClustering component generates well-separated clusters. 
\begin{equation}
\label{eq:sil}
\textit{Silhouette Score} = \frac{b - a}{\max(a, b)}
\end{equation}

\textbf{Calinski-Harabasz Index$\big\uparrow$}~\cite{CalinskiHarabasz}: 
The \textit{Calinski-Harabasz Index} measures the ratio of between-cluster dispersion to within-cluster dispersion. In Equation~\ref{eq:ch}, \(\text{Tr}(B_k)\) is the trace of the between-cluster dispersion matrix, \(\text{Tr}(W_k)\) is the trace of the within-cluster dispersion matrix, \(k\) is the number of clusters, and \(n\) is the total number of entities. This index is also known as the Variance Ratio Criterion, and higher values indicate that the between-cluster variance is much greater than the within-cluster variance.

\begin{equation}
\label{eq:ch}
\textit{Calinski-Harabasz Index} = \frac{\text{Tr}(B_k)/(k-1)}{\text{Tr}(W_k)/(n-k)}
\end{equation}

\textbf{Davies-Bouldin Index$\big\downarrow$}~\cite{DaviesBouldin}: 
The \textit{Davies-Bouldin Index} quantifies the average similarity between clusters by comparing the distance between clusters relative to the sizes of the clusters. The lower values for this index indicate better clustering quality and a higher distance between clusters. In Equation~\ref{eq:db}, \(\sigma_i\) denotes the dispersion within cluster \(i\), \(d(c_i, c_j)\) is the distance between the centroids of clusters \(i\) and \(j\), and \(k\) is the number of clusters.

\begin{equation}
\label{eq:db}
\textit{Davies-Bouldin Index} = \frac{1}{k}\sum_{i=1}^{k}\max_{j \neq i}\left(\frac{\sigma_i + \sigma_j}{d(c_i, c_j)}\right)
\end{equation}

Among the components of the \EnrichEvent framework, the \TrendingDataExtraction module detects and filters irrelevant tweets from incoming message blocks. 
To evaluate its performance, we measure \textit{Accuracy}, \textit{Recall}, and \textit{Precision} in a supervised setting. 
Trending data estimation from social streams is inherently imbalanced, as most tweets contain general daily updates rather than trending information and discussing events. Therefore, in addition to \textit{macro} and \textit{micro} averages, we report label-specific performance of the \TrendingDataExtraction module based on the \textit{Accuracy}, \textit{Recall}, and \textit{Precision} outcomes. 
To quantify the impact of the \TrendingDataExtraction module on the scalability of the \EnrichEvent framework, we compare the \textit{Runtime} of the system with and without this module enabled. 
Simulation results indicate that an optimally performing \TrendingDataExtraction module significantly reduces the framework's \textit{Runtime} while maintaining acceptable levels of \textit{Accuracy}, \textit{Recall}, and \textit{Precision}.

\subsection{Implementation Settings}
We implement the \EnrichEvent framework and replicate the baselines using a Google Colab machine equipped with a Tesla T4 GPU and 12GB of RAM. 
While replicating the baselines, we set a similar setting across all the frameworks to eliminate any variation from the implementation settings, and the variation in outcome measures will only result from the design and performance of frameworks.
The replication package for the \EnrichEvent framework is publicly available on Github~\cite{replication2023EnrichEvent}. Researchers can use this replication package to reproduce our results and future research. 

Table~\ref{tab:hyper} presents the details of the hyperparameters used in this study.
The $\sigma$ hyperparameter controls the sensitivity of the \TrendingDataExtraction component in filtering out irrelevant tweets. To achieve moderate sensitivity in classification, we set $\sigma$ to 0.3 across all frameworks. This means that tweets are removed from the incoming message block if their probability of referring to an event falls below this threshold.
The $\beta$ hyperparameter determines the number of selected tweets per entity when generating the embedding vector in the \ContextualKnowledgeEnhancement component. We set $\beta$ to 4 for all baselines, except for ~\textit{Fedoryszak et al.}~\cite{RealTime_EventDetection_Twitter} and ~\textit{Yang et al.}~\cite{Yang2018AnED}, which do not capture the contextual representation of tweets. For these two frameworks, we set $\beta$ to 0.
The \ContextualKnowledgeEnhancement component constructs the entity-to-entity matrix based on the occurrence and embedding matrices. The $\alpha$ hyperparameter balances the contribution of these two matrices. Since ~\textit{Fedoryszak et al.}~\cite{RealTime_EventDetection_Twitter} and ~\textit{Yang et al.}~\cite{Yang2018AnED} do not generate an embedding matrix, we set $\alpha$ to 0 for them. We set the $\alpha$ hyperparameter to 0.5 for all other frameworks to balance structural and contextual entity representations.
The $\delta$ hyperparameter defines the minimum number of common entities required to establish an edge between nodes in the bipartite graph within the \EventChainFormation component. To reduce noise caused by frequently mentioned entities, we set $\delta$ to 3 across all frameworks.

\begin{table}
\begin{center}
\caption{The values of hyperparameters are selected based on trial and error to ensure stable performance for all the studied frameworks.
}
\label{tab:hyper}
\resizebox{\columnwidth}{!}{
\begin{tabular}{|P{2.2cm}|P{6.3cm}|P{0.6cm}|}
\hline
\rule{0pt}{8pt}
\textbf{Hyperparameter}    & \textbf{Description} &  \textbf{Value}  \\

\hline
\rule{0pt}{9pt}
        $\sigma$        & Minimum threshold for the probability of referring to an event in the Trend Detection component            &   $0.3$     \\

\hline
\rule{0pt}{9pt}
        $\beta$         & Total number of selected tweets for generating the embedding vectors for each entity       &   $4$    \\

\hline
\rule{0pt}{9pt}
        $\alpha$        & Weight of Embedding matrix while building the entity-to-entity distance matrix         &   $0.5$     \\

\hline
\rule{0pt}{9pt}
       $\delta$        & Minimum number of common entities to add an edge among clusters in the \EventChainFormation component    &   $3$   \\ 
\hline


\end{tabular}}
\end{center}
\vspace{-10pt}
\end{table}

\section{Experiments and Results} \label{sec:results}

In contrast to prior works~\cite{Yang2018AnED, RealTime_EventDetection_Twitter} that try to detect unspecified events from structural and lexical aspects of tweets, our novel contribution is to leverage the contextual knowledge of tweets.
We aim to leverage this knowledge to represent users' diverse perspectives and opinions about events.
In the first RQ, replicate the baselines from prior works and compare their performance with our novel \EnrichEvent framework.
In the second RQ, we examine the quality of generated entity clusters from different perspectives.
In addition to efficacy, the event detection frameworks should have high scalability, so in RQ 3, we examine the impact of \TrendingDataExtraction on the scalability and performance of the \EnrichEvent framework.
Although we calculate all the outcome measures mentioned in Section~\ref{sec:metric}, in this study, we are most interested in \Consolidation and \Discrimination outcomes.
An ideal unspecified event detection framework achieves a high \Consolidation and low \Discrimination, which shows that the framework achieves a good understanding of events and distinguishes them perfectly.

\subsection{RQ1: Replication \& Evaluation}

\begin{table}
\centering
{\caption{
The \Consolidation outcome shows how the existing entities in event chains refer to a single event and how the existing entities in each cluster are related together. On the other hand, the \Discrimination outcome demonstrates how well the frameworks discriminate the events from each other and measures how well they avoid grouping unrelated entities together.
}
\label{tab:result}}
\begin{tabular}{|c|c|c|}

\hline
\rule{0pt}{12pt}
\textbf{Method} & \textbf{Consolidation$\big\uparrow$} &\textbf{Discrimination$\big\downarrow$}
\\
\hline
\rule{0pt}{9pt}
\textit{Word2vec}~\cite{word2vec}& 63.46\% & 28.94\% \\
\hline
\rule{0pt}{9pt}
\textit{BERT}~\cite{BERT}& 67.01\%& 15.97\% \\
\hline
\rule{0pt}{9pt}
\textit{Yang et al.}~\cite{Yang2018AnED}& 38.42\%& 37.64\% \\
\hline
\rule{0pt}{9pt}
\textit{Fedoryszak et al.}~\cite{RealTime_EventDetection_Twitter}& 65.20\%& 25.90\% \\
\hline
\rule{0pt}{9pt}
\EnrichEvent & 87.41\% & 10.00\% \\
\hline
\end{tabular}
\vspace{-10pt}
\end{table}

\textit{How well does the EnrichEvent framework perform compared to the baselines?}

When users discuss social events online, they debate various aspects of an event influenced by their background and perspective. 
Previous studies~\cite{Yang2018AnED, RealTime_EventDetection_Twitter} focus on detecting events by analyzing the structural and lexical patterns in users' discussions.
In contrast, this study enriches tweet representations by emphasizing the contextual aspects of events.
Our proposed framework, \EnrichEvent, integrates contextual information with lexical and structural features to capture the diverse perspectives in social media discussions.

In this research question, we replicate the baselines and compare their performance with our novel framework.
Then, we feed the message blocks to the frameworks and wait to identify the occurring events. 
Based on their definition, the \Consolidation and \Discrimination outcomes are calculated for a single event.
Furthermore, we separately calculate the \Consolidation and \Discrimination outcomes for each detected event. We also report the average values across all events as the final results for the studied frameworks in Table~\ref{tab:result}.

The results in Table~\ref{tab:result} demonstrate that the \EnrichEvent framework outperforms all the baselines in both \Consolidation and \Discrimination outcomes.
Specifically, \EnrichEvent achieves an average \Consolidation of 87.41\% and an average \Discrimination of 10.00\% across all events.
These findings indicate that our framework distinguishes events effectively and ensures that most selected entities are relevant.
The Word2vec and BERT baselines, which primarily focus on contextual representation, attain average \Consolidation values of 63.46\% and 67.01\% and average \Discrimination values of 28.94\% and 15.97\%, respectively.
The framework by Fedoryszak et al.~\cite{RealTime_EventDetection_Twitter}, which leverages structural and lexical features, achieves moderate results with averages of 65.20\% for \Consolidation and 25.90\% for \Discrimination.
Notably, the approach by Yang et al.~\cite{Yang2018AnED} records the lowest \Consolidation (38.42\%) and the highest \Discrimination (37.64\%), underscoring its limitations in effectively detecting occurring events.

\begin{tcolorbox}
Simultaneously integrating the contextual, structural, and lexical knowledge in tweets is crucial for capturing the diverse perspectives of users and ensuring robust event detection. Reliance on any aspect is inadequate and does not yield considerable results.
\end{tcolorbox}

\subsection{RQ2: Representation Enhancement}

\begin{figure}
    \centering
    \includegraphics[scale=0.28]{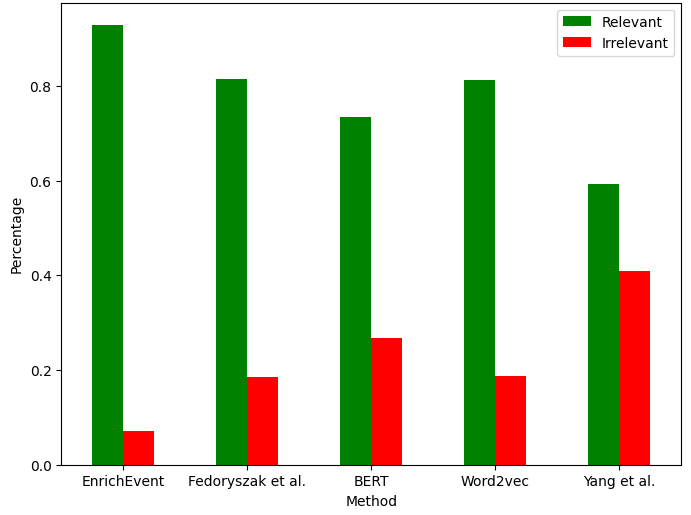}
    \caption{In this experiment, we average the \textit{Fraction of Relevant Entities (FRE)} across all the involved clusters in event chains. The higher values show that the \EventClustering module clusters the entities more effectively. The \EnrichEvent framework leverages the lexical, structural, and contextual knowledge simultaneously, and compared to baselines, it reaches a better perspective about the users' opinion about the ongoing events.}
    \label{fig:fraction}
\end{figure}

\begin{figure}
    \centering
    \includegraphics[scale=0.25]{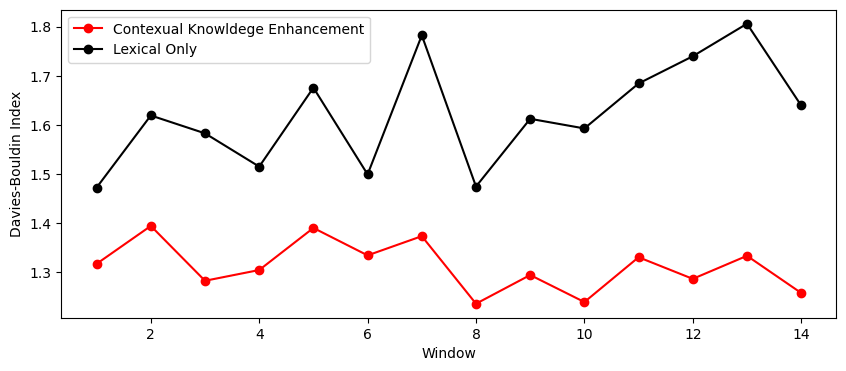}
    \caption{While leveraging the contextual knowledge, the \EventClustering module generates less overlap and scattered clusters, and their centroids are far from each other.}
    \label{fig:davies}
\end{figure}

\textit{How does leveraging the contextual knowledge along with the lexical and structural aspects enhance the quality of event detection?}

In this study, we define an event as a continuous sequence (chain) of clusters derived from incoming message blocks, where each cluster groups together entities referring to the same occurrence. 
Within this framework, it is crucial that the entities within each cluster consistently pertain to the same event, while distinct events are clearly differentiated.
Our previous findings demonstrated that incorporating the contextual knowledge of entities enriches their representation, thereby enhancing the event detection process.
Building on this, the current research question explores the impact of contextual knowledge on both the representation of entities and the overall quality of clusters.

In our first experiment, we calculate the \textit{Fraction of Relevant Entities (FRE)} across all clusters within the event chains, comparing outcomes across various baselines. The results in Figure~\ref{fig:fraction} reveal that the \EnrichEvent, \textit{Fedoryszak et al.}, \textit{Word2evc}, \textit{BERT}, and \textit{Yang et al.} frameworks achieve FRE values of 92.89\%, 81.49\%, 81.28\%, 79.95\%, and 59.16\%, respectively.
Notably, while the \textit{Fedoryszak et al.} framework relies solely on co-occurrence vectors of \NamedEntities, the \EnrichEvent framework leverages contextual knowledge to achieve a markedly higher FRE, underscoring a more refined and comprehensive understanding of events.
This indicates that the \EnrichEvent framework effectively captures event-specific information, ensuring that most entities within each cluster are relevant.

\begin{figure}
    \centering
    \includegraphics[scale=0.25]{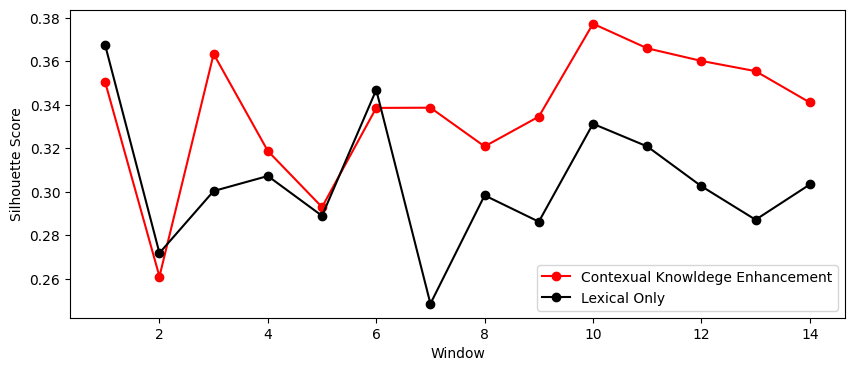}
    \caption{In this experiment, we calculate the \textit{Silhouette Score} for the generated clusters for each message block and compare it to the scenario in which we only settle for the existing lexical and structural knowledge in tweets. The results demonstrate that when we add the contextual knowledge, the entities in each cluster become more similar to each other.}
    \label{fig:silhouette}
\end{figure}

\begin{figure}
    \centering
    \includegraphics[scale=0.25]{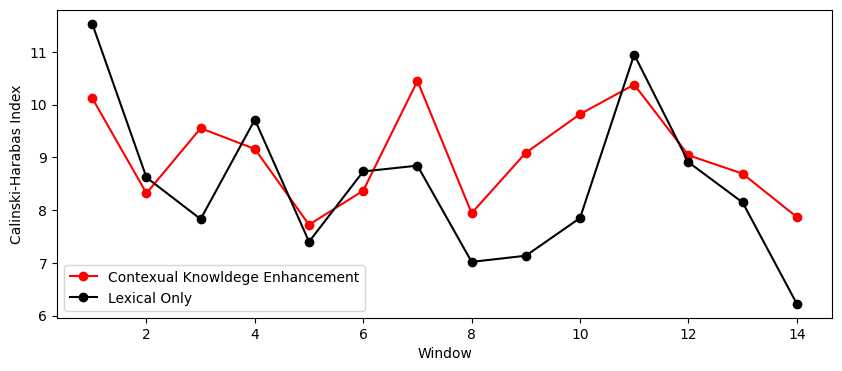}
    \caption{The results show that in most message blocks, leveraging the contextual knowledge leads to more distinct clusters and less dispersion within each cluster.}
    \label{fig:calinski}
\end{figure}

The following experiment examines the influence of contextual knowledge on cluster quality.
For this purpose, we compute the \textit{Silhouette}, \textit{Calinski-Harabasz}, and \textit{Davies-Bouldin} indices for clusters generated by the \EventClustering module, which organizes entities based on their combined lexical, structural, and contextual representations. We then compare these results with clusters formed using only lexical and structural knowledge.
The \textit{Silhouette Score}, \textit{Calinski-Harabasz Index}, and \textit{Davies-Bouldin Index} serve as metrics for assessing cluster quality by measuring intra-cluster similarity, inter-cluster variance, and distances between clusters, respectively.
As shown in Figure~\ref{fig:davies}, the integration of contextual knowledge increases the separation between clusters by 10\% to 38\%.
Moreover, the enhancements observed in the \textit{Silhouette Score} and \textit{Calinski-Harabasz Index} in Figures~\ref{fig:silhouette} and~\ref{fig:calinski} further confirm that enriching entity representations with contextual knowledge significantly improves cluster quality across most message blocks.

\begin{tcolorbox}
Integrating contextual knowledge enriches entity representation and substantially enhances cluster quality, leading to a more precise and discriminative event detection process.
\end{tcolorbox}

\subsection{RQ3: Impact of Trending Data Extraction Module}

\begin{table}
\begin{center}
\caption{Label-specific performance of the \TrendingDataExtraction module on the test data. To ensure that tweets with a high probability of referring to potential events are not missed, we prioritize the Recall outcome for Label-1 among the classification metrics.}
\label{tab:trend}
\begin{tabular}{|c|c|c|c|c|}
\hline
\rule{0pt}{12pt}
&\textbf{Precision} &\textbf{Recall} & \textbf{F1-Score} & \textbf{Support}
\\
\hline
\rule{0pt}{9pt}
\textbf{Label-0}& 0.97& 0.90& 0.93 & 312254\\
\textbf{Label-1}& 0.53& 0.78& 0.63 & 45307\\
\hline
\rule{0pt}{9pt}
\textbf{Macro-Average}& 0.75& 0.84& 0.78 & 357561 \\
\textbf{Weighted-Average}& 0.91 & 0.89& 0.89 & 357561 \\
\hline
\end{tabular}
\end{center}
\vspace{-15pt}
\end{table}

\textit{What is the impact of filtering out the pointless tweets in message blocks on the performance of the EnrichEvent framework?}

Most daily tweets consist of personal updates rather than discussions about trending events. Prior studies~\cite{critien2022bitcoin, cavalli2021cnn} highlight the importance of extracting meaningful, event-related data from Twitter.
The \TrendingDataExtraction component within the \EnrichEvent framework is designed to eliminate non-informative data and enhance both the effectiveness and scalability of the framework.
This research question evaluates the efficiency of the \TrendingDataExtraction module and its impact on the performance of the \EnrichEvent framework.

In the first experiment, we assess the performance of the \TrendingDataExtraction module using classification metrics.
Table~\ref{tab:trend} presents label-specific results, demonstrating that the module effectively detects pointless tweets that do not correspond to potential events.
For Label-0, which represents non-event-related tweets, the classifier achieves a \textit{Precision}, \textit{Recall}, and \textit{F1-Score} of 0.97, 0.90, and 0.93, respectively.
On the other hand, it is critical to filter the tweets with high \textit{true positive rate (Recall)} to ensure that the model does not remove the event-related tweets mistakenly. We emphasize the \textit{Recall} metric since we want to filter tweets with great confidence, and we prefer to see all tweets with label one as much as possible. For Label-1, which corresponds to newsworthy tweets, the \TrendingDataExtraction module achieves a \textit{Precision}, \textit{Recall}, and \textit{F1-Score} of 0.53, 0.78, and 0.63, respectively. 
Although learning discriminative patterns is inherently challenging in imbalanced datasets, the \textit{Macro-Average} scores of 0.75 for \textit{Precision}, 0.84 for \textit{Recall}, and 0.78 for \textit{F1-Score} indicate that the \TrendingDataExtraction module effectively filters out pointless data while preserving a reasonable recall for event-related tweets.

Scalability is crucial for any event detection framework, as low scalability makes real-time event detection impractical.
To measure the impact of the \TrendingDataExtraction module on scalability, we disable the component and monitor the pipeline \textit{Runtime}.
Disabling this module leads to a 50\% increase in \textit{Runtime}, highlighting its significant role in improving the framework's efficiency.

\begin{tcolorbox}
Filtering non-informative tweets significantly improves the efficiency and scalability of event detection frameworks while maintaining high-quality event detection. 
\end{tcolorbox}

\section{Discussion} \label{sec:discussion}
In this section, we discuss our findings from the results and the study's limitations.

\subsection{Results and Findings}

In this study, we evaluate the performance of event detection frameworks based on \Consolidation and \Discrimination outcomes.
The \Consolidation outcome quantifies the quality of event chains by calculating how well all chain entities refer to the same event.
Conversely, the \Discrimination metric evaluates how effectively a framework distinguishes one event from another by avoiding grouping unrelated entities.  
Intuitively, we can think of an algorithm putting all entities in a single cluster as achieving 100\% \Consolidation but 0\% \Discrimination. On the other hand, creating a cluster for each entity will yield 0\% \Consolidation and 100\% \Discrimination.

An ideal event detection framework balances high \Consolidation and low \Discrimination outcomes.
Comparing the results of \textit{Fedoryszak et al.}~\cite{RealTime_EventDetection_Twitter} and \textit{BERT}~\cite{BERT} demonstrates that exploiting the contextual knowledge reduces the \Discrimination outcome by 10\% (25.9\% vs. 15.9\%). Still, it does not individually affect the value of the \Consolidation (65.2\% vs. 67\%). These results show that contextual knowledge assists the framework in achieving a better understanding of users' opinions and distinguishing events from each other more perfectly.
Compared to \textit{BERT}~\cite{BERT}, the \EnrichEvent framework enhances the \Consolidation outcome from 67\% to 87.4\%. This result shows that integrating the contextual, structural, and lexical knowledge in tweets is crucial for capturing the diverse perspectives of users and ensuring robust event detection.
From the results of \textit{Yang et al.}~\cite{Yang2018AnED}, we see that hashtags are not good representatives of tweets and can not present users' opinions about events. Since users usually post hashtags to specify the topic of tweets, they do not transfer considerable knowledge for event detection.

In the pipeline of the \EnrichEvent framework, the main goal of the \TrendingDataExtraction module is to enhance the scalability and help the \ContextualKnowledgeEnhancement to enrich representation matrices more effectively by removing the pointless tweets.
In addition to scalability, the \TrendingDataExtraction can be used to customize the event detection process. 
More specifically, embedding a domain-specific classifier in the \TrendingDataExtraction module makes it possible to identify the events in that specific domain. 
For instance, if you only want to detect sports-related events, it is possible to easily adjust the \TrendingDataExtraction module and identify the tweets in which users share their opinions about the sports-related events.

\subsection{Generalization}

We collect real social streams from Twitter over two weeks to run our experiments.
Since we leverage pre-trained language models in the components, one of the main advantages of the \EnrichEvent framework is that it is possible to easily adjust the language models for various languages and other streaming data sources like news~\cite{lin2024multi}, Wikipedia~\cite{SEDTW}, etc.
Future work is necessary to validate our results in other contexts.
We also provide a replication package~\cite{replication2023EnrichEvent} for future researchers who wish to replicate or extend our results.

\subsection{Limitations}

In this study, we assume that each occurrence of a \textit{Named Entity} or \textit{Hashtag} corresponds to a single event.  
This assumption enables us to cluster related entities together and construct coherent event chains.  
Moreover, we apply the \textit{Hungarian algorithm} to match nodes in bipartite graphs, ensuring maximum flow while constructing these chains.  
However, in real-world settings, a single cluster may encompass multiple events, and there can be multiple ideal matchings for building event chains.  
Future work may explore incremental clustering approaches~\cite{ozdikis2017incremental, cao2021knowledge, guo2024unsupervised, cao2024hierarchical}, community detection models~\cite{singh2024event}, and time-varying hypergraphs~\cite{yan2023abnormal} to overcome these limitations.  

Prior studies utilize benchmarks such as \textit{Event2012}~\cite{Event2012} and \textit{Event2018}~\cite{Event2018} to evaluate event detection frameworks and compare their performance with baseline methods.  
These benchmarks are accessible via the Twitter API. However, we could not collect data directly from it due to budget constraints.  
Therefore, we gathered 76K tweets directly from Twitter using the \textit{Scrapy} tool.  
Owing to resource constraints for manual labeling, we under-sample the data from 76K to 7K tweets (500 tweets per message block) to conduct our experiments.  
We sample tweets from each message block based on token frequency to maintain the dataset's distribution. However, this under-sampling may affect the generalizability of our findings.  
Furthermore, since prior literature does not provide a labeled dataset to train the classifier in the \TrendingDataExtraction component, we collect 1.6 million tweets across different categories and automatically label them using domain-specific key phrases.  
Although we define diverse categories and precise key phrases, this data collection and labeling approach may limit the overall generalization of our results.  

Although the \TrendingDataExtraction component enhances the scalability and efficiency of the \EnrichEvent framework, its overall performance depends on the accuracy of this component.  
Our results in Table~\ref{tab:trend} validate the efficacy of this module.  
However, the component incorrectly filters out 15\% of tweets referring to potential events.  
Moreover, the \textit{Hungarian algorithm} that we embed in the \EventChainFormation component has a time complexity of $O(n^3)$, which limits scalability.  
Identifying more efficient alternatives for both the \textit{Hungarian algorithm} and the \TrendingDataExtraction component is a promising direction for future work.

\section{Conclusion} \label{sec:conclusion}

While discussing an ongoing event, users share different opinions and personal perspectives based on their unique experiences, backgrounds, and knowledge. 
In contrast to prior works~\cite{Yang2018AnED, RealTime_EventDetection_Twitter} that settle for the lexical and structural patterns for identification of \textit{unspecified events}, the \EnrichEvent framework also leverages the contextual knowledge to represent users' diverse opinions and discussions about ongoing events.
The following findings highlight the potential of contextual knowledge to represent users' personal perspectives and opinions about social events.
\begin{itemize}
    \item 
    Using the contextual, structural, and lexical knowledge in tweets simultaneously is crucial to represent the users' diverse perspectives about social events. However, relying on any aspect is inadequate and does not ensure robust event detection.
    
    \item 
    Leveraging the contextual knowledge enriches the representation of events and enhances the quality of clusters based on their variance, similarity, and distance. 
    
    \item 
    In most tweets, users post a general update about their daily routines, and the event detection frameworks need to filter out pointless tweets to ensure their scalability.

\end{itemize}

The recent advancements in Large Language Models (LLMs) have demonstrated their ability to process and understand natural language with remarkable sophistication~\cite{leas2024using}. 
These models offer significant potential for efficiently analyzing large volumes of streaming data and automatically extracting structured knowledge graphs~\cite{shiri2024decompose}.
Furthermore, LLMs can improve the representation of users' perspectives on ongoing social events by capturing nuanced opinions and contextual meanings.
Exploring LLMs for event representation learning and user opinion modeling presents a compelling avenue for future research.

On social media platforms, users express their opinions about ongoing events.
Applying sentiment analysis techniques on the detected social events is an interesting future direction to understand the emotional tone of individual messages and the overall event.
This analysis helps reveal public sentiment, distinguishing between satisfied and dissatisfied users while capturing broader social reactions~\cite{golchin2025emotion}.
Examining sentiment trends over time can also provide deeper insights into how users perceive and engage with significant events on online social networks~\cite{petrescu2024edsa}.

\bibliography{IEEE-Transactions-LaTeX2e-templates-and-instructions/EnrichEvent_ref}
\bibliographystyle{IEEEtran}

\appendices

\section{EnrichEvent Output} \label{Appendix:output}
\noindent After detecting the events, the \EnrichEvent framework returns a JSON file for each event as an output, including detailed information about the detected events, which is listed in Table~\ref{tab:output}. This information contains the summary of each event, the period in which each event occurred, and detailed data about the hashtags, users, and named entities that referred to each event. 

\begin{table}[!h]
\begin{center}
\caption{The extracted information about each event in a JSON file that the \textit{Storage} module saves in the database as the final output of \EnrichEvent for each event.}
\label{tab:output}
\resizebox{\columnwidth}{!}{
\begin{tabular}{|P{3.2cm}|P{6.3cm}|}
\hline
\rule{0pt}{8pt} \textbf{Key} & \textbf{Description} \\ \hline
\rule{0pt}{9pt} Event ID & A unique identifier assigned to the event. \\ \hline
\rule{0pt}{9pt} Event Period & The period during which the event occurs. \\ \hline
\rule{0pt}{9pt} Event Summary & A brief description of the event. \\ \hline
\rule{0pt}{9pt} Tweet with Most Likes & The tweet that received the highest number of likes. \\ \hline
\rule{0pt}{9pt} Tweet with Most Retweets & The tweet that was retweeted the most. \\ \hline
\rule{0pt}{9pt} Tweet with Most Replies & The tweet with the highest number of replies. \\ \hline
\rule{0pt}{9pt} Count Unique Hashtags & The total number of distinct hashtags. \\ \hline
\rule{0pt}{9pt} Count Total Hashtags & The aggregate number of all hashtags used. \\ \hline
\rule{0pt}{9pt} Most Frequent Hashtags & The hashtags that appear most frequently in the event. \\ \hline
\rule{0pt}{9pt} Count Unique Users & The total count of distinct users who discuss the event. \\ \hline
\rule{0pt}{9pt} Most Frequent Users & The users who posted the most tweets. \\ \hline
\rule{0pt}{9pt} Count Unique Entities & The count of unique entities corresponds to the event. \\ \hline
\rule{0pt}{9pt} Count Total Entities & The aggregate count of all entities mentioned. \\ \hline
\rule{0pt}{9pt} Most Frequent Entities & The entities mentioned most frequently in the event. \\ \hline
\rule{0pt}{9pt} Word Cloud & A visual representation of the most common words related to the event. \\ \hline
\end{tabular}}
\end{center}
\vspace{-10pt}
\end{table}

\section{Datasets Details} \label{Appendix:data}
\noindent\textbf{Labeling of Trend Detection Dataset}: To prepare a generalized trend detection dataset, we defined 12 categories for tweets' topics. Furthermore, we explored various accounts and collected all the tweets posted by experts and active users in each category using the Twitter API. We leveraged domain-specific key phrases to label the collected tweets in each category. Specifically, we defined a collection of key phrases for each category and checked their occurrence within the tweets to label the dataset. We consider labeling one for the tweets even if one of the key phrases appears in them.

\begin{table}
\begin{center}
\caption{Proposed categories for tweets' topics. We aim to collect adequate tweets in each category and train the \TrendingDataExtraction module on diverse fields.}
\label{tab:categories}
\resizebox{\columnwidth}{!}{
\begin{tabular}{|c|c|c|c|c|c|}
\hline
\rule{0pt}{9pt} Sports & Technology & Art & Industry & Celebrities & Video Games\\
\hline
\rule{0pt}{9pt} Crypto & Politics & Health & Environment & Economic & Social  \\
\hline
\end{tabular}}
\end{center}
\end{table}

\noindent\textbf{Event Detection Dataset}: In contrast to the Trend Detection Dataset, we did not consider any user specification for the Event Detection Dataset, and we only limited the language of the tweets to Persian.

\end{document}